# A Greedy Part Assignment Algorithm for Realtime Multi-Person 2D Pose Estimation

Srenivas Varadarajan
Intel Labs, Intel Corporation
Bangalore, India.
srenivas.varadarajan@intel.com

Parual Datta
Intel Labs, Intel Corporation
Bangalore, India.
parual.datta@intel.com

Omesh Tickoo
Intel Labs, Intel Corporation
Hillsboro, OR, USA.
omesh.tickoo@intel.com

## Abstract

*Human pose-estimation in a multi-person image involves detection of various body parts and grouping them into individual person clusters. While the former task is challenging due to mutual occlusions, the combinatorial complexity of the latter task is very high. We propose a greedy part assignment algorithm that exploits the inherent structure of the human body to achieve a lower complexity, compared to any of the prior published works. This is accomplished by (i) reducing the number of part-candidates using the estimated number of people in the image, (ii) doing a greedy sequential assignment of part-classes, following the kinematic chain from head to ankle (iii) doing a greedy assignment of parts in each part-class set, to person-clusters (iv) limiting the candidate person clusters to the most proximal clusters using human anthropometric data and (v) using only a specific subset of pre-assigned parts for establishing pairwise structural constraints. We show that, these steps sparsify the body-parts relationship graph and reduces the algorithm's complexity to be linear in the number of candidates of any single part-class. We also propose methods for improving the accuracy of pose-estimation by (i) spawning person-clusters from any unassigned significant body part and (ii) suppressing hallucinated parts. On the MPII multi-person pose database, pose-estimation using the proposed method takes only 0.14 seconds per image. We show that, our proposed algorithm, by using a large spatial and structural context, achieves the state-of-the-art accuracy on both MPII and WAF multi-person pose datasets, demonstrating the robustness of our approach.*

## 1. Introduction

Human pose estimation is an important building block for performing several computer vision tasks like action recognition [23] and human-object interaction recognition [24] and computing object affordance [25]. Human pose refers to the configuration of human body parts in a 3D space or a 2D image. Single-person 2D pose-estimation methods employ a set of hand-crafted features [16] or learnt features from a deep neural network (DNN) [17-18], to model the appearance and configuration of human body parts of an individual. We address the problem of multi-person 2D pose estimation in this work. It involves 3 sub-problems, namely, (i) Part Detection, which infers the pixel locations of all the body parts (ii) Part Labelling, which assigns a body part class (eg. head, neck, elbow) to each body part location (iii) Part Assignment, which involves grouping the labelled body parts into person-clusters. The major challenge for multi-person pose estimation in the wild, stems from a higher potential for persons to fully or partially occlude the body parts of each other in the captured image. These mutual occlusions along with articulated nature of human body parts, makes it very hard for a single model to capture all the possible body-part appearances and spatial configurations in a multi-person image. Some of the simplest of approaches like [19-20], which perform human detection followed by single-person pose-estimation around each person, lead to huge inaccuracies in multi-person pose estimation. Some of the recent approaches [1-3] get an initial set of part candidates using DNNs and solve the part-labelling and part-assignment problems jointly using Integer Linear Programming (ILP). But ILP is NP hard and it becomes intractable as the number of part-candidates increases.

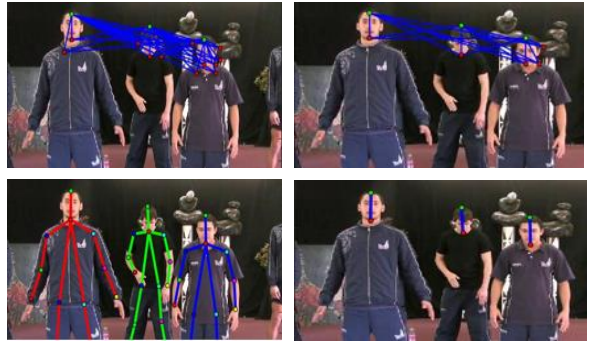

Figure 1. An illustration of the proposed algorithm: Clockwise from top-left: significant head detections (green dots) and neck candidates (red dots), neck detections after clustering, partial person-clusters after neck assignment, final person-clusters after performing candidate clustering and part-assignment for all parts.

In this paper, we propose a multi-person pose estimation algorithm that completely avoids the combinatorial complexity of ILP and exploits the inherent structure of the human body to greedily solve an otherwise generic graph-partitioning problem. We reduce the part-candidates by estimating the number of people in the image. Although, locating people to reduce computational complexity seems similar to the works in [3], [9] and [11], we do not perform part-assignment in local regions around each detected person, but jointly assign the parts of all persons in a large spatial context. Although, we do a sequential part assignment similar to [7], we use a larger structural context of a subset of pre-assigned parts rather than using only the immediate predecessors as in [7].

Our algorithm focuses on reducing the combinatorial complexity of the part-assignment problem in 5 ways. First, the number of candidates of each part class is reduced to the approximate number of people in the image through a clustering step. Secondly, we do a greedy sequentially assignment of body parts to people clusters, one part-class at a time moving progressively down the kinematic chain. Since the number of parts at a single stage are much smaller than the total number of parts, this incremental approach helps in reducing the complexity by a huge factor. Thirdly, in each stage, we greedily assign one part at a time without considering other part-assignments. Fourthly, we consider only the most proximal candidate person-clusters while assigning the considered part. We resort to using human anthropometric data for accomplishing this. Lastly, in the part-assignment step, we reduce the number of pairwise constraints between the body parts by using only a subset of previously assigned parts in the partial person-clusters. All these steps help in introducing a high degree of sparsity in the graph partitioning problem introduced in [1], as illustrated in Fig 1. As shown in Section 3, this sparsity enables our algorithm to achieve a complexity which is linear in number of candidates of any single part-class. This is much lower than the exponential time complexity of ILP based algorithms [1-3] or the cubic time complexity of the Hungarian algorithm [8] employed in [7]. Due to the complexity reduction, the proposed algorithm takes only 0.14 seconds per image on the MPII multi-person dataset[6]. Even though our algorithm greedily assigns one part at a time, by using the considered part's association with a large structural and spatial context, it achieves the state-of-art accuracy of 72.6% on the MPII pose dataset [6]. We also show that our algorithm achieves state-of-the-art results on the WAF pose dataset [13].

## 2. Related Work

The recent approaches for multi-person pose estimation use a DNN for detecting the body parts and a graphical network for assigning them to individuals. The DeepCut algorithm [1] first obtains a set of part-confidence maps using an adapted Fast R-CNN[22]. The candidates obtained from these maps, are jointly labelled and partitioned into individual poses using an ILP solver, by establishing constraints on the mutually exclusive labelling of parts and pairwise relationship of parts assigned to an individual. But ILP is a NP hard problem, exhibiting an exponential time complexity in the number of part-candidates. In Deeper Cut [2], a better part detection is performed using a deep Residual Network (ResNet-101) [4] and incorporating image-conditioned pairwise terms. The running time of ILP is reduced by stronger probabilistic models for capturing the pairwise probabilities and an incremental optimization strategy. This achieves a speed up of 2-3 orders of magnitude compared to DeepCut [1]. An approach aimed at reducing the complexity of the ILP through local joint assignment is proposed in [3]. Convolution pose machines [15] is used in [3] for obtaining the part-confidence score maps. The part-labelling and part-assignment problems are solved through multiple ILPs, one ILP evaluated for each human bounding box region. This approach achieves a 50 times speed up compared to DeeperCut [2] at the cost of poor accuracy. Another similar work, which first detects humans using SSD detector [26] followed by single-person pose-estimation at each human location is proposed in [11]. The approach uses a Stacked-hourglass network (SHN) [17] for computing single person pose. The initial person proposals are improved using a spatial transformation network before estimating the pose and mapped back to the image using a de-transform network. Training the SHN and the two transformation networks jointly, results in a better accuracy when compared to [3]. More recently, Part Affinity Fields (PAFs) that signify the limb probabilities between every pair of connected part-types is proposed in [7]. PAFs are learnt using a multi-stage CNN. The person-clustering problem is formulated as partitioning a K-Partite graph into several connected components, where K is the number of part-classes. The PAFs are used as edge-weights of this graph. The problem is solved using a series of K bipartite graph matching problems, each solved using the Hungarian algorithm [8]. In [9], the graph partitioning problem formulated in DeeperCut [2] is solved quickly using a modified KL algorithm [10], instead of ILP. Similar to our approach, [9] reduces the complexity of the graph-partitioning problem by considering only a subset of pairwise associations between different body parts. The work also proposes a top-down approach which solves a set of simplified graph-partitioning problems in local regions around the head-detections. The pairwise probabilities of parts, conditioned on the head location is inferred from a CNN. The approach in [5] uses a pair of DNNs for every body part-class, one for part-candidate detections and another for group-assignments for each detection. The CNNs are based on the SHN architecture, proposed in [17]. The significant necks 'anchor' person-clusters and all parts with matching tags that fall within a prescribed distance from the anchor points, are assigned to that cluster. This approach is very fast as it uses DNNs for part-assignments rather than solving a graphical network.

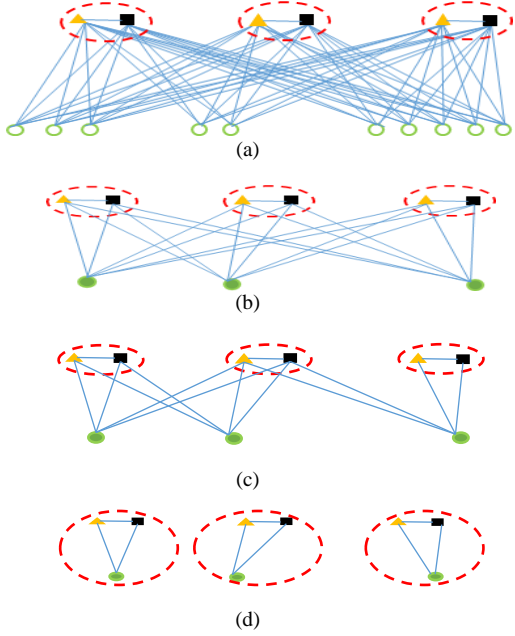

Figure 2. A visualization of the sparsification of the body-part relationship graph by the proposed algorithm, during the assignment of right shoulder (Rshr) to $N_H=3$, partial person clusters, each containing a pre-assigned head and a neck. The heads, necks and Rshr are indicated by orange triangles, black squares and green circles, respectively. The pre-assigned parts in the partial person clusters are enclosed in red dotted circles. (a) graph at the start of Rshr assignment (j=3) with $N_j = 10$ part-candidates. (b) graph after clustering the part-candidates as Rshr parts (c) graph after selecting the proximal candidate clusters for each Rshr part (d) graph after the assignment of each Rshr part to one of the person clusters.

## 3. Proposed Method

We first formulate the problem of multi-person pose-estimation and elucidate how the different steps of our proposed algorithm help in reducing the complexity of the part-assignment problem. We then explain the details of our algorithm. We assume that a set of confidence maps for each body part class and a set of part-association maps can be computed using an a priori modelled set of DNNs, similar to those used in [2] and [7]. A part confidence map gives the probability of occurrence of the corresponding part-class at every pixel location while an association map gives the probability of co-occurrence for every pair of part candidates, in the same person.

**Problem Formulation and Complexity Analysis**

We first obtain a set of body part candidates, $D_j$ of each part-class j by the non-maximal suppression of the confidence map of part j, where $D_j = \{d_j^i : j \in \{1,2,....J\} \text{ and } i \in \{1,2...N_j\}\}$, where $N_j$ is the number of candidates of part-class j, $d_j^i$ represents the $i^{th}$ candidate of the $j^{th}$ part class and J =14, is the total number of part classes. Let $P_{ij}$ denote the unary probability of $d_j^i$ while $P_{ljmk}$ denote the co-occurrence probability of $d_j^l$ and $d_k^m$ in the same person. The multi-person pose estimation problem can be viewed as retaining a subset of all candidate parts from all part-classes and assigning each part to one of the $N_H$ person clusters, $\beta = \{\beta_h : h \in \{1,2,....N_H\}\}$, while satisfying the constraint that not more than one part of any part-class is assigned to the same person cluster. The problem can now be visualized as a J-Partite graph in which the nodes represent the part-candidates and the edge-weights reflect the pairwise association probabilities. There are a total of $N_D$ nodes in the graph, where $N_D = \sum_{j=1}^{J} N_j$. The solution to the part-assignment problem amounts to partitioning this graph into $N_H$ disjoint subgraphs such that, each subgraph represents a person-cluster. This solution can be represented using a set of indicator variables $Z_{i,j,h} \in (0,1)$ which capture the assignment of the $i^{th}$ candidate of the $j^{th}$ part-class to the $h^{th}$ person cluster. To begin with, $N_H$ is unknown in the considered image. Our algorithm begins by estimating $N_H$ from the number of significant head detections. Each head location initializes a person cluster and at this stage, $Z_{i,j,h} = 1$ for all permissible combinations of i, j and h. The body parts are assigned to these person clusters greedily, considering one part-class at a time, moving sequentially down the kinematic chain from neck to ankle. At each step j (corresponding to the $j^{th}$ part-class) along the kinematic chain, two steps are performed:

**Step 1**: The body part-class candidate-set $D_j$ is first spatially clustered to $N_H$ clusters through K-means clustering with 100 iterations. This step has a complexity of $O(N_H N_j)$. As a result of clustering, we obtain $C_j = \{d_j^c : c \in \{1,2,....N_H\}\}$ part-cluster centers that denote the final body parts of the part-class j. The candidate-part clustering sets a subset of indicator variables to zero as shown below:

$$Z_{i,j,h} = \begin{cases} 1 & d_j^i \in C_j \\ 0 & otherwise \end{cases} \quad (1)$$

**Step 2**: Each of the body parts $d_j^c$ is assigned to a partial person-cluster, $h$, which has the maximum cluster affinity score with that part. The cluster affinity score, $\pi_{c,j,h}$ between a part, $d_j^c$ and a person-cluster, $\beta_h$, is computed as the average pairwise probability of $d_j^c$ with the $T$ prior assigned parts of $h$. Since $|C_j| = N_H$ and $|\beta| = N_H$, this step incurs a complexity of $O(TN_H^2)$. As a result of part-assignment step another major set of indicator variables are set to zero as shown below:

$$Z_{i,j,h} = \begin{cases} 1 & d_j^i \in C_j \text{ and } h = \underset{t}{\text{argmax}}(\pi_{c,j,t}) \\ 0 & otherwise \end{cases} \quad (2)$$

As we move down the kinematic chain, the number of prior assigned parts, $T$, in each partial person-cluster increases progressively. In order to keep the complexity constant we use only a subset of $L$ parts as predecessors while assigning the current part. In images with large number of people (large $N_H$), the complexity of part-assignment is high. In order to reduce the complexity of the part assignment step only a subset of $M$ most proximal person-clusters are considered for part assignment. We use the human anthropometric data and scale of the image to compute $M$ adaptively. So the complexity of Step 2 reduces to $O(MN_H L)$. The overall complexity of the proposed algorithm is $O(N_H N_j)+O(MN_H L)$. Since the number of part candidates, $N_j$ are much larger than any other parameter, our algorithm's complexity is linear in the maximum number of part candidates belonging to any part-class. All the prior approaches based on ILP are NP-hard in complexity. While DeepCut[1] is NP-hard in $N_D^2 \cdot J^2$, DeeperCut[2] is NP-hard in $N_D^2$ due to its incremental optimization. The local joint assignment proposed in [3] converts the original problem into $N_H$ sub-problems each solved using ILP. Each ILP is NP-hard in $N_{D_h}^2$, where $N_{D_h}$ is the number of part-candidates falling in the h[th] person's bounding box. Finally, the approach in [7] relaxes the original J-Partite graph partitioning problem into a series of J bi-partite graph matching problems, each solving the association of adjacent part-class candidates in the kinematic chain. The Hungarian- algorithm [8] used at each stage, has a complexity of $O(N_j^3)$. The $O(N_j)$ complexity of our proposed algorithm is hence very much lower than that of any of the previously suggested graph-based approaches.

An alternate way for expressing the complexity reduction achieved by our algorithm, is through the visualization of the body-parts relationship graph as illustrated in Fig 2. As shown in Fig 2, during the assignment of the right shoulder candidates (j =3) to a set of $N_H = 3$, partial person-clusters, each step of the proposed algorithm progressively sparisifies the graph. This sparsity reduces the complexity and hence speeds up the multi-person part assignment problem.

**Multi-person Part-Assignment**
The flow-chart of the proposed method is outlined in Fig 3. We use the DeeperCut [2] models in our work for obtaining the part confidence and pairwise association maps of body parts. The sequential assignment starts from head because heads are the most reliable human part to detect in images [2]. We do non-maximal suppression on the head-confidence map and retain only the significant local maxima, whose unary probabilities exceed 0.5. The number of people in the image, $N_H$, is initially estimated as the number of significant heads, each of which seeds a person-cluster. Other body parts are added to these person clusters sequentially, moving down the kinematic chain from neck to ankle. The following steps are repeated for each part-class, starting from neck (j=2).

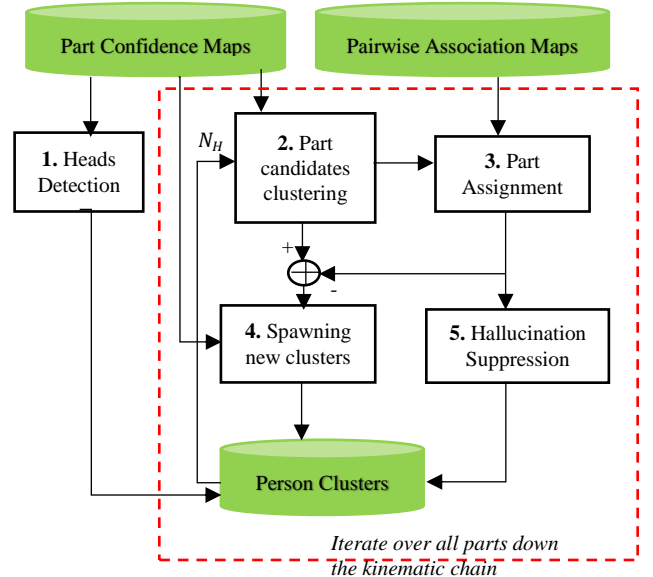

Figure 3. Flow-chart of the proposed algorithm.

### 3.1 Candidate Parts Clustering
The number of part-candidates, $d_j^i$, even after non-maximal suppression on the part-confidence maps is much larger than the actual number of body parts in the image. As the complexity of sequential part assignment to person-clusters increases linearly as a function of part candidates ($N_j$), it is important to reduce the number of part candidates. As it is natural to expect one part per part-class per person in a scene with no occluded body parts, we reduce the number of part-candidates to $N_H$ through a clustering algorithm. A spatial K-means clustering is performed on all the part candidates of a particular part-class where, $K = N_H + 2$. The two additional cluster centers permitted at each step is to accommodate parts belonging to partially visible people. The candidates nearest to the center of each cluster are qualified as final body parts, $d_j^c$ for the part assignment step. If there are multiple closest members to a cluster center, then the candidate with the highest unary probability is labelled as the body-part.

### 3.2 Sequential Part Assignment
Once a set of body parts are selected in the previous step, they have to be assigned to the correct partial person-clusters. Each of these partial clusters include parts that were already assigned prior to the considered part-assignment. While assigning a body part, $d_j^c$, of part-class $j$ to one of the considered person-clusters, $\beta_h$, the pairwise probabilities between $d_j^c$, and the pre-assigned parts of $\beta_h$, are used to compute a cluster affinity score, $\pi_{c,j,h}$, for that part with respect to the candidate person cluster.

$$\pi_{c,j,h} = \frac{1}{|\beta_h|}\sum_{d' \in \beta_h} P_{d_j d' t} \qquad (3)$$

| Body Part | Normalized Distance from top of head ($\alpha$) |
|---|---|
| Chin | 0.130 |
| Neck | 0.182 |
| Shoulders | 0.224 |
| Elbow | 0.410 |
| Wrist | 0.556 |
| Hips | 0.481 |
| Knees | 0.726 |
| Ankles | 0.972 |

Table 1. Distances of body parts from top of head normalized w.r.t. body height derived from the anthropometric data in [12].

where $d'$ is a pre-assigned part of class $t$, in the candidate person cluster $\beta_h$ and $|\beta_h|$ is the cardinality of β. The body part is assigned to the cluster, $\beta_{h*}$, with the maximum cluster affinity score, $\pi_{c,j,h*}$ and denoted as follows:

$$\beta_{h*} = \beta_{h*} U \{d_j^c\} \quad (4)$$

The complexity of the part assignment is further reduced by the following steps.

**(i) Using a subset of candidate person-clusters:**
While assigning a considered part to a partial cluster, the likelihood of the part getting assigned to a distant cluster is low. In order to select the most proximal candidate-clusters, we estimate the scale of the image from the average head length, $y$, expressed in pixels. Further, we use the human body part ratios given in [12], to compute the expected distance of a part from its corresponding head location as follows:

$$R = \frac{y}{0.13} \alpha \quad (5)$$

where $\alpha$ is the distance of the part from the top of the head (normalized w.r.t human height) and enclosed in Table 1. The maximum possible displacement of a body part from the corresponding head is given by $R$ and this happens when the human body frame or the plane of articulation is parallel to the image plane of the camera. In order to account for inaccuracies in scale estimation, all heads in a radius of 1.5R from the current part are considered for assignment. Since computing the head length requires the assignment of neck parts to the correct heads, this speed up is used only for all other parts starting from shoulders. As shown in Fig 4, the proximal clusters for the assignment of right hip of 6th person and right elbow of 3rd person (from left) are enclosed by green and red circles, respectively.

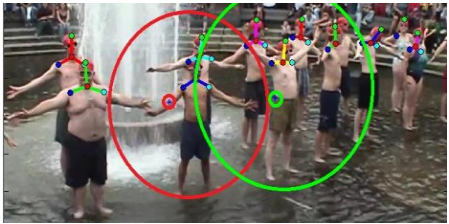

Figure 4. An illustration of Candidate Person-Cluster Selection. The two proximal clusters for the assignment of right elbow of the 3rd person from left are enclosed by the large red circle while the five proximal clusters for the assignment of the right hip of the 6th person from left are enclosed by the large green circle.

| Part-class | | Predecessor |
|---|---|---|
| j | Name | |
| 2 | Neck | Head |
| 3 | R Shr | Head, Neck |
| 4 | L Shr | Head, Neck, R Shr |
| 5 | R Elbow | Head, Neck, R Shr |
| 6 | L Elbow | Head, Neck, L Shr |
| 7 | R Wrist | Head, Neck, R Shr, R Elbow |
| 8 | L Wrist | Head, Neck, L Shr, L Elbow |
| 9 | R Hip | Head, Neck, L Shr, R Shr |
| 10 | L Hip | Head, Neck, R Shr, L Shr |
| 11 | R Knee | Head, Neck, R Shr, L Shr, R Hip |
| 12 | L Knee | Head, Neck, R Shr, L Shr, L Hip |
| 13 | R Ankle | R Hip, R Knee |
| 14 | L Ankle | L Hip, L Knee |

Table 2. Predecessors for body part-classes in the Kinematic chain, used during part assignment to partial person clusters.

**(ii) Using a subset of prior-assigned parts**
We propose using only a specific sub-set of previously assigned parts as "Predecessors" for part assignment in (3), while still preserving the structural context. The predecessors for the various parts are enclosed in Table 2. We choose the prescribed set of predecessors to accommodate profile views, where only one side of the human body is visible. As the head and neck are the most reliably detected body parts, they are used as predecessors to all other upper body parts.

### 3.3 Spawning new person clusters
Our algorithm draws its efficiency from limiting body parts to approximately the number of persons in the image. Only relying on significant head-detections to estimate the people-count, can prevent the detection of all parts of a person-cluster if its head is occluded. As mentioned in Section 3.1, in order to accommodate people with occluded heads, the candidates are clustered into $K = N_H + 2$ clusters and one member from each cluster is qualified as a body part of the corresponding part-class. Of these $N_H$ body parts are assigned to one of the $N_H$ partial person clusters, as described in Section 3.2. Now, the two unassigned body parts can be either real, belonging to people with occluded heads or can be spurious, arising purely due to our clustering mechanism. As shown in the flow-chart (Fig. 3), each unassigned body part, $d_j^u$ is further analyzed. Each $d_j^u$ having a significant unary probability of 0.35 or higher, is inferred as a real body part with all occluded predecessors and a new person-cluster, $\beta_s$ is spawned as follows.

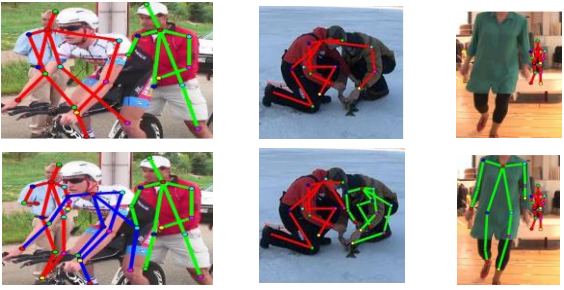

Figure 5. Examples of missed person-clusters due to undetected heads (top row) and results after spawning new clusters based on other significant body parts.

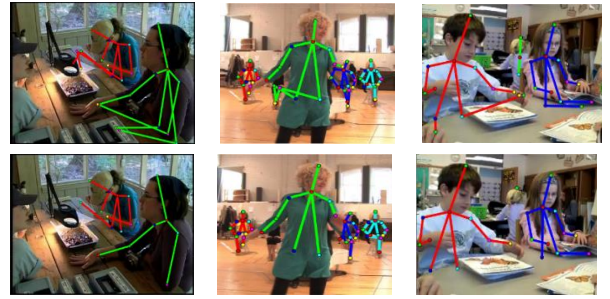

Figure 6. Examples of images with hallucinated parts (top row) and results after hallucination suppression (bottom row).

$$\beta_s = \{d_j^u\} \quad \text{where } s = N_H + 1 \quad (6)$$
$$N_H = N_H + 1 \quad (7)$$

Spawning new clusters based on other significant body parts, helps in reducing the sensitivity of the proposed algorithm to undetected heads and improves the accuracy of our algorithm as illustrated through Fig 5.

### 3.4 Suppression of Hallucinated Parts

As mentioned in Section 3.1, the number of visible parts of each part-class are estimated to be approximately the number of people in the image. But this need not be true in images with occluded body parts. In such cases, spurious parts get detected at locations where that part is not truly present and assigned to partial person-clusters. For suppressing the hallucinated parts [15], a structural-probability score, $S_c$, for each body part, $d_j^c$ is computed as follows.

$$S_c = \frac{1}{2}(P_{cj} + \pi_{c,j,h*}) \quad (8)$$

where $\pi_{c,j,h*}$ is the maximum of the cluster affinity scores computed in (3). Only the parts having a significant structural probability score of 0.6 are retained while others are suppressed. Suppression of hallucinated parts is illustrated through some examples in Fig 6. Images with hallucinated body parts are shown in the top row of Fig 6 while the corresponding results after hallucination suppression are shown in the bottom row of Fig 6.

## 4. Results and Analysis

The efficacy of the proposed algorithm to achieve the best possible combination of speed and accuracy is shown by evaluating its performance on two most popular multi-person human pose estimation datasets, namely, (i) MPII Multi-Person Human Pose Dataset [6] and (ii) We are a Family (WAF) Dataset [13] and comparing its performance with some of the state-of- the-art methods. Accuracy is established through the average precision (AP) metric as evaluated in the previous works [1-3]. The time taken by an implementation of the proposed part-assignment algorithm in Matlab R2015b, on a system with Intel Core i7 CPU clocking at 1200 MHz with 64 GB RAM and 16MB L2 cache, is enclosed here. The time taken by the compared state-of the art works on similar systems are directly obtained from their published results.

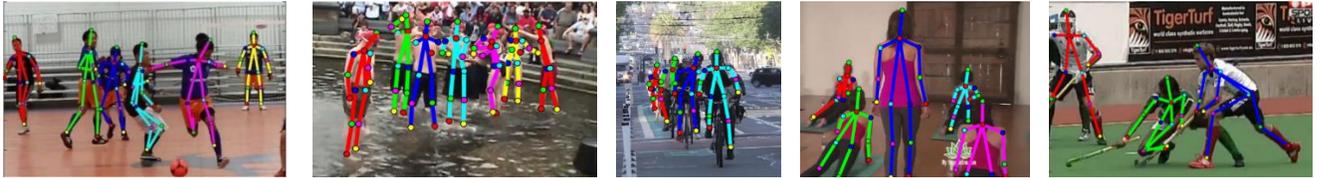

Figure 7. Examples of successful pose estimation results on applying the proposed algorithm.

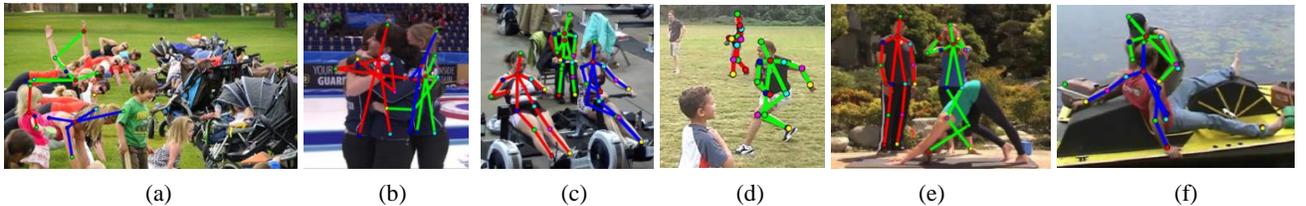

| (a) | (b) | (c) | (d) | (e) | (f) |

Figure 8. Some failure cases on applying the proposed algorithm: (a) and (b) missed persons due to severe occlusions, (c) and (d) double-counting (in which both left and right feet fall on the same location) errors, (e) and (f) missed parts due to rare poses.

| Method | Head | Shoulder | Elbow | Wrist | Hip | Knee | Ankle | mAP | Time (sec) |
|---|---|---|---|---|---|---|---|---|---|
| full-testing Set | | | | | | | | | |
| Cao et al. [7] | 91.2 | 87.6 | 77.7 | 66.8 | 75.4 | 68.9 | 61.7 | 75.6 | 1.24 |
| Fang et al. [11] | 87.7 | 85.1 | 77.3 | 68.5 | 73.5 | 70.9 | 63.8 | 75.3 | 0.8 |
| Insafutdinov et al.[9] | 88.8 | 87 | 75.9 | 64.9 | 74.2 | 68.8 | 60.5 | 74.3 | 0.005 |
| Newell&Deng [5] | 87.3 | 82.4 | 70.4 | 58.8 | 68.6 | 62.3 | 55.7 | 69.4 | 0.06 |
| DeeperCut [2] | 78.4 | 72.5 | 60.2 | 51 | 57.2 | 52 | 45.4 | 59.5 | 485 |
| Iqbal et al [3] | 58.4 | 53.9 | 44.5 | 35 | 42.2 | 36.7 | 31.1 | 43.1 | 10 |
| **Our model** | **92.1** | **85.9** | **72.9** | **61.6** | **72** | **64.6** | **56.6** | **72.2** | **0.14** |
| subset of 288 Images as in [1] | | | | | | | | | |
| Cao et al. [7] | 92.9 | 91.3 | 82.3 | 72.6 | 76.0 | 70.9 | 66.8 | 79 | 1.1 |
| Insafutdinov et al.[9] | 92.2 | 91.3 | 80.8 | 71.4 | 79.1 | 72.6 | 67.8 | 79.3 | 0.005 |
| Fang et al. [11] | 89.4 | 88.5 | 81 | 75.4 | 73.7 | 75.4 | 66.5 | 78.6 | 0.75 |
| Newell&Deng [5] | 91.5 | 87.2 | 75.9 | 65.4 | 72.2 | 67.0 | 62.1 | 74.5 | 0.06 |
| DeeperCut [2] | 87.9 | 84 | 71.9 | 63.9 | 68.8 | 63.8 | 58.1 | 71.2 | 230 |
| Iqbal et al [3] | 70 | 65.2 | 56.4 | 46.1 | 52.7 | 47.9 | 44.5 | 54.7 | 10 |
| DeepCut [1] | 73.1 | 71.7 | 58 | 39.9 | 56.1 | 43.5 | 31.9 | 53.5 | 57995 |
| **Our model** | **92.9** | **88.8** | **77.7** | **67.8** | **74.6** | **67** | **63.8** | **76.1** | **0.12** |

Table 3. Pose-estimation results (AP) on the full MPII Multi-Person Dataset [6].

**MPII Multi-person Dataset**: This dataset consists of 1758 test images containing groups of people. A group is a region of the image which contains overlapping human body parts of two or more people. The number of people in each group varies from 2 to 12. The evaluation results for the full set of 1758 images and on the subset of 288 images used in DeepCut[1] are shown in Table 3. The proposed algorithm achieves an overall accuracy of 72.2% on the full data-set and takes an average of 0.14 seconds per group. This includes the time for obtaining the unary and pairwise probabilities using the DNN in [2] and for the proposed greedy part assignment algorithm. On the reduced dataset of 288 images, it achieves a 76% accuracy. Our method is more accurate and faster than any of the prior approaches [1-3] that use ILP for part assignment. When compared to the more recent approaches [5, 7, 9, 11], the proposed algorithm is the third fastest method using only a sub-optimal implementation of part-assignment in Python. The proposed algorithm also achieves the state-of-the-art on this dataset [6]. Some of the successful pose-estimation results of the proposed algorithm are shown in Fig 7, while some of the failure cases are shown in Fig 8. There are 3 types of pose-estimation errors that occur due to the application of the proposed algorithm. These are (i) completely missed persons due to severe occlusions, (ii) double-counting errors due to the similarity in the appearance of left and right body parts (eg. knees and ankles) and (iii) missed body parts due to rare poses, as shown in Fig 8. In case of severe occlusions, the unary part probabilities are very much low and hence get suppressed. In case of rare poses, the pairwise probabilities of visible body parts are insignificant as those configurations were not well-represented in the training data. In a total of 103 erroneous detections observed in the 1758 images of the MPII test dataset, there were 33 cases of completely missed persons due to severe occlusions, 26 cases of double-counting and 44 cases of missed body parts due to rare poses.

**Ablation Analysis**: We evaluate the contributions from different steps of the proposed algorithm by experimenting with a validation set of 1000 images from the MPII multi-person pose dataset and enclose the results in Table 4. We take as baseline configuration, just the candidate part assignment step of the proposed algorithm. The candidate parts are assigned to person clusters seeded by significant heads. In the absence of candidate part-clustering, more than one candidate of a part-class gets assigned to a person-cluster. In order to ensure that each person-cluster has not more than one part per class, at the end of each stage, for each partial cluster, only the candidate with the maximum cluster affinity score is retained. The accuracy and the average time for this baseline configuration is shown in the first row of Table 4. On adding candidate clustering to the baseline algorithm, the average time per image drastically reduces by 66% w.r.t baseline at the cost of 1% decrease in mAP. Adding the proximal person cluster selection step, does not further impact the mAP or the speed as shown in the second row. This is because the Validation dataset did not contain images with large number of people and hence the complexity reduction due to this step is insignificant. Using only a subset of predecessors for part-assignment results in a further speed up of 22% without impacting accuracy as shown in the fourth row of Table 4. As shown in the fifth row, spawning of new clusters increases the accuracy by 1.2% with no additional cost on speed. Finally, hallucination suppression improves the accuracy by 0.3% on this dataset without affecting the speed of pose-estimation. The improvements in mAP from the last 2 steps can be much higher on datasets with a higher degree of body-part occlusions. Overall, our algorithm results in a speed up of 4 times and a slight improvement in accuracy compared to the baseline, on this validation dataset.

| Configuration | mAP | Time(s) |
|---|---|---|
| 1. Baseline | 79.6 | 0.537 |
| 2. + Candidate Clustering | 78.6 | 0.182 |
| 3. + Proximal Clusters | 78.6 | 0.175 |
| 4. + Subset of Predecessors | 78.6 | 0.142 |
| 5. + Spawning new clusters | 79.8 | 0.144 |
| 6. + Hallucination Suppression | 80.1 | 0.147 |

Table 4. Results of the ablation experiments of the proposed algorithm on accuracy and speed, evaluated on a validation dataset of 1000 images from MPII Multi-person pose database [6].

| Method | Head | U Arms | L Arms | Torso | mPCP | Time (s) |
|---|---|---|---|---|---|---|
| Eichner& Ferrari [13] | 97.6 | 68.2 | 48.1 | 86.1 | 69.4 | - |
| Chen& Yuille[14] | 98.5 | 77.2 | 71.3 | 88.5 | 80.7 | - |
| DeepCut[1] | 99.3 | 81.5 | 79.5 | 87.1 | 84.7 | 22000 |
| DeeperCut[2] | 99.3 | 83.8 | 81.9 | 87.1 | 86.3 | 13 |
| our method | **100** | **86.3** | **83.7** | **97** | **89.5** | **0.18** |

Table 5. Performance on the upper body parts for visible Torsos on We Are a Family (WAF) dataset [13].

**We Are a Family (WAF) dataset:** The performance of the proposed algorithm on the WAF dataset evaluated using the original implementation in [13], is enclosed in Table 5. As shown in Table 5, mPCP is the metric used for comparing the performance of various algorithms. mPCP is the Percentage of Correctly detected Parts (PCP) computed only on people whose upper body positions (head and trunk) match the ground-truth [2]. The proposed algorithm achieves the highest mPCP of 89.5% for the various upper body parts from head to hips. The mPCPs of other state-of-the-art approaches for pose-estimation are enclosed in Table 5. The average time per image of the proposed algorithm on WAF dataset (Table 5) is slightly higher than that of MPII Multi-person dataset (Table 3) because the average number of people in each group is higher in the WAF dataset. Some of the successful and failure cases for pose-estimation using the proposed method on the WAF dataset are shown in Fig 9.

A demo video showing the real-time multi-person pose-estimation, using the proposed algorithm, is enclosed as a part of the supplementary material.

## 5. Conclusion

In this paper, we propose a greedy part assignment algorithm for multi-person pose estimation, which quickly qualifies a subset of part-candidates and groups them to sets of person-clusters. Our algorithm exploits the inherent structure present in the human body to sparsify the body-parts relationship graph by (i) reducing the number of part-candidates, (ii) greedily assigning part-classes to person-clusters (iii) greedily assigning part-candidates with in each part-class to person-clusters and (iv) selecting a set of most proximal person-clusters using human anthropometric data. We show that these steps reduce the complexity of our algorithm to be linear in the number of candidates of any single part-class. In spite of the complexity reduction, we preserve the accuracy of multi-person pose estimation by using a large structural and a spatial context. We demonstrate the ability of our algorithm to simultaneously attain the state-of-the-art accuracy and a huge reduction in complexity, by evaluating its performance on two popular multi-person pose datasets.

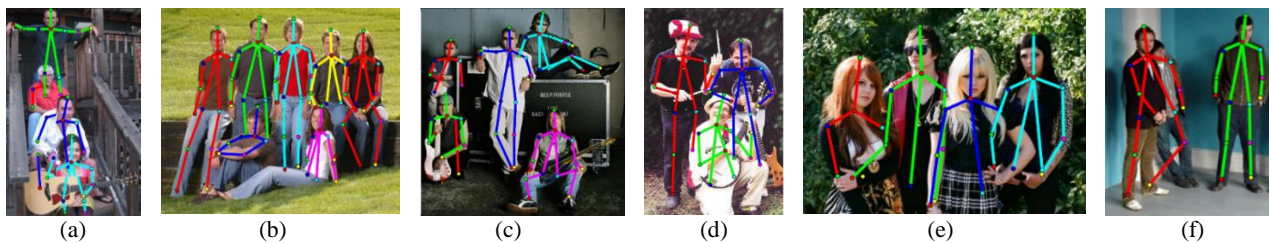

(a)     (b)     (c)     (d)     (e)     (f)

Figure 9. Some results of the proposed algorithm on We Are a Family dataset [7]. Some successful results are shown in images (a) – (d). Some failures due to incorrect assignment of parts to people are shown in (e)-(f)